\documentclass[10pt,twocolumn]{IEEEtran}
\usepackage[utf8]{inputenc}
\usepackage{amsmath}
\usepackage{graphicx}
\usepackage[numbers]{natbib}  
\usepackage{hyperref}
\usepackage{lipsum} 

\title{Precision Knowledge Editing: Enhancing Safety in Large Language Models}

\author{
    \IEEEauthorblockN{
        Xuying Li,
        Zhuo Li,
        Yuji Kosuga,
        Yasuhiro Yoshida,
        and Victor Bian
    }
    \IEEEauthorblockA{
        HydroX AI\\
        Email: \{xuyingl, zhuoli, yujikosuga, yasuhiro, victor\}@hydrox.ai
    }
}

\begin{document}

\maketitle

\begin{abstract}
Large language models (LLMs) have demonstrated remarkable capabilities, but they also pose risks related to the generation of toxic or harmful content. This work introduces Precision Knowledge Editing (PKE), an advanced technique that builds upon existing knowledge editing methods to more effectively identify and modify toxic parameter regions within LLMs. By leveraging neuron weight tracking and activation pathway tracing, PKE achieves finer granularity in toxic content management compared to previous methods like Detoxifying Instance Neuron Modification (DINM). Our experiments demonstrate that PKE significantly reduces the attack success rate (ASR) across various models, including Llama2-7b and Llama-3-8b-instruct, while maintaining overall model performance. Additionally, we also compared the performance of some closed-source models (gpt-4-0613 and Claude 3 Sonnet) in our experiments, and found that models adjusted using our method far outperformed the closed-source models in terms of safety. This research contributes to the ongoing efforts to make LLMs safer and more reliable for real-world applications.
\end{abstract}


\section{Introduction}
The rapid advancement of large language models (LLMs) has revolutionized natural language processing, enabling unprecedented capabilities in text generation, comprehension, and task completion. However, these powerful models also present significant challenges, particularly in managing the potential generation of toxic, biased, or harmful content. This issue has become a critical concern for researchers and practitioners alike, as the deployment of LLMs in real-world applications requires robust safety measures.

Existing approaches to address this problem include knowledge editing techniques, which aim to modify specific pieces of information or correct undesirable behaviors in LLMs without retraining the entire model. One prominent method in this field is Detoxifying Instance Neuron Modification (DINM), which has shown promise in efficiently identifying and modifying toxic regions in LLMs through single test instances. However, our research has uncovered limitations in the DINM approach, particularly in its effectiveness across different model architectures.

The primary finding of our study indicates that the DINM method is ineffective for some models due to the small gap between harmful ($h_{harm}$) and safe ($h_{safe}$) parameters. This ineffectiveness may stem from prompt design issues, but more likely from data-related problems during model training. For instance, if the training data doesn't sufficiently distinguish between safe and harmful outputs, the model's internal representations of these concepts may be too similar, making it difficult for DINM to effectively separate them.

To address these limitations, we propose Precision Knowledge Editing (PKE), a novel approach that allows for more accurate identification and modification of toxic parameters within LLMs. PKE builds upon the foundations laid by DINM but incorporates more sophisticated techniques for parameter analysis and modification. By tracking changes in neuron weights and tracing activation pathways, PKE achieves a finer granularity in identifying toxic regions, leading to more targeted and effective edits.

Our approach not only improves the robustness of knowledge editing but also provides a more scalable solution for managing toxic content in LLMs. Through extensive experimentation and analysis, we demonstrate that PKE outperforms DINM in reducing the attack success rate (ASR) across multiple model architectures, while maintaining the models' overall performance on general tasks.

\section{Preliminaries}
\subsection{Knowledge Editing in LLMs}
Knowledge editing refers to the process of modifying specific pieces of information or correcting undesirable behaviors in large language models (LLMs) without retraining the entire model. This field has gained significant attention as LLMs, such as GPT-3 and its successors, are increasingly deployed in real-world applications. The ability to alter or remove toxic content, outdated information, or biased outputs is critical for maintaining the safety and reliability of these models.

Existing techniques in knowledge editing include approaches like ROME (Rank-One Model Editing) and MEND (Modular Editing Networks with Gradients), which enable fast and localized adjustments to model parameters without sacrificing overall performance. However, most of these techniques focus on factual knowledge correction, leaving the challenge of toxic content largely unaddressed.

Detoxifying Instance Neuron Modification (DINM) has emerged as a prominent solution to this problem by targeting specific neurons responsible for toxic outputs. Despite its success, DINM's simplified approach often lacks the granularity required to handle more complex toxic content in various contexts. Our work extends this line of research by proposing Precision Knowledge Editing (PKE), a more nuanced technique that tracks neuron weight changes and activation pathways to edit toxic regions at a finer level of granularity.

\subsection{Challenges in Toxic Content Identification}
Identifying toxic content within LLMs presents multiple challenges, primarily due to the context-dependent nature of toxicity and the subtle variations in prompts that can trigger harmful outputs. Unlike factual errors, where incorrect knowledge can be directly identified, toxic behavior may not be immediately apparent until the model generates harmful or biased outputs in response to specific queries.

Another major challenge lies in distinguishing between "safe" and "harmful" parameters within the model. In our research, we found that merely measuring the gap between harmful parameters ($h_{harm}$) and safe parameters ($h_{safe}$) is insufficient for accurate identification. This is because the relationship between model parameters and output toxicity is often non-linear and context-dependent.

\section{Method}
In this section, we describe the proposed Precision Knowledge Editing (PKE) method, which enhances the accuracy of knowledge editing by focusing on identifying and modifying toxic parameter regions within large language models (LLMs). PKE builds upon the existing Detoxifying Instance Neuron Modification (DINM) technique but introduces a series of improvements to achieve more precise and effective edits. The following subsections outline the mathematical formulations and the editing process in detail.

\subsection{Mathematical Formulations}
To effectively edit toxic parameters in LLMs, PKE introduces more advanced mathematical techniques to identify key regions for editing. These formulations allow PKE to track fine-grained changes within the model, enabling targeted edits.

\subsubsection{Neuron Weight Change}
One of the primary metrics used in PKE is the change in neuron weights. Let the weight matrix of the $l$-th layer be $W_l$, the input vector be $x$, and the output vector be $y$. The goal is to track how the weight matrix evolves before and after applying PKE, which is computed as:
\[
\Delta W_l = \| W_l^{\text{after}} - W_l^{\text{before}} \|_F
\]
where $\| \cdot \|_F$ represents the Frobenius norm. The Frobenius norm is chosen for its ability to measure the overall magnitude of the weight changes, providing insight into how much the network has adjusted its internal parameters during knowledge editing.

\textit{Motivation}: By calculating the change in weight across multiple layers, we can identify neurons that have undergone significant modifications. These neurons are likely contributing to the generation of toxic content and are prime candidates for further edits.

\subsubsection{Activation Path Tracking}
The gradient of the activation path represents the sensitivity of the model’s output to changes in the weights of specific layers. The gradient of the $l$-th layer is computed as:
\[
g_l = \frac{\partial \hat{y}}{\partial W_l}
\]
where $\hat{y}$ is the predicted output. The change in the activation path can then be determined by the norm of this gradient:
\[
\Delta g_l = \left\| \frac{\partial \hat{y}}{\partial W_l} \right\|
\]

\textit{Motivation}: Tracking the gradient of the activation path helps pinpoint the layers that have the most influence on the model's toxic behavior. Layers with significant gradient changes are potential "hot spots" where toxic behavior can be mitigated with minimal impact on the model's overall performance.

\subsubsection{Average Toxicity Across Multiple Instances}
To ensure the edits generalize across multiple test instances, PKE computes the average toxicity score across several samples. For a set of test instances $x_i$, the model produces output $\hat{y}_i$, and we assign a toxicity score $T(\hat{y}_i)$ to each output. The average toxicity in the $l$-th layer is then calculated as:
\[
\overline{T}_l = \frac{1}{N} \sum_{i=1}^{N} T(\hat{y}_i^{(l)})
\]
where $N$ is the number of test instances, and $\hat{y}_i^{(l)}$ represents the model's output after the $l$-th layer. This step ensures that the knowledge editing process reduces toxic content across diverse inputs, not just for individual cases.

\textit{Motivation}: By averaging the toxicity scores, we can avoid overfitting the edits to specific test cases. This formulation ensures that the model’s improvements are robust and apply to a broader range of inputs, thereby improving the model's general safety.

\subsubsection{Local Region Identification}
One of the central objectives of PKE is to identify local regions in the model that contribute disproportionately to toxic content generation. This is done by computing the change in activation for individual neurons. The change in activation for the $j$-th neuron in the $l$-th layer is computed as:
\[
\Delta h_{l,j} = \left| h_{l,j}^{\text{after}} - h_{l,j}^{\text{before}} \right|
\]
where $h_{l,j}$ is the activation of neuron $j$ in layer $l$. The total activation change for the layer is then computed as:
\[
\Delta H_l = \sum_{j=1}^{m_l} \Delta h_{l,j}
\]
where $m_l$ is the number of neurons in layer $l$. This allows PKE to focus on the most problematic neurons within a given layer.

\textit{Motivation}: By isolating and adjusting these neurons, PKE ensures that the edits are highly localized, reducing the risk of overcorrecting and compromising the model’s overall performance.

\subsubsection{Custom Loss Function}
The custom loss function used in PKE is designed to balance two objectives: reducing toxicity and maintaining the correctness of the model’s outputs. The total loss is represented as:
\[
L = \alpha T(\hat{y}) + \beta C(\hat{y})
\]
where $T(\hat{y})$ represents the toxicity score of the model’s output, and $C(\hat{y})$ represents the correctness score. $\alpha$ and $\beta$ are weight coefficients that determine the balance between the two goals.

\textit{Motivation}: This custom loss function ensures that as PKE reduces toxicity, it does not inadvertently degrade the overall performance or correctness of the model.

\subsection{Precision Knowledge Editing (PKE) Process}
PKE follows a structured process to identify, target, and modify the toxic parameters within an LLM. The process can be summarized in the following steps:

\textbf{Step 1: Identify Key Layers.} The first step involves identifying the key layers responsible for toxic content generation by tracking the activation path. The layer $l^*$ with the highest activation gradient is selected as the primary target:
\[
l^* = \arg\max_l \Delta g_l
\]
This ensures that the edits focus on the layers that have the most significant influence on the model's outputs.

\textbf{Step 2: Pinpoint Critical Neurons.} Once the key layer is identified, PKE targets the neurons within that layer that have undergone the most significant weight changes. The neuron $j^*$ with the largest weight change is selected for editing:
\[
j^* = \arg\max_j \Delta h_{l^*,j}
\]
This step helps localize the edits to the most problematic neurons, minimizing the risk of affecting unrelated parts of the model.

\textbf{Step 3: Apply Knowledge Editing.} After selecting the critical neuron, PKE modifies the weight matrix for the selected neuron to reduce its contribution to toxic outputs. The weight update is applied using the following rule:
\[
W_{l^*, j^*}^{\text{new}} = W_{l^*, j^*}^{\text{old}} - \eta \cdot \Delta W_{l^*, j^*}
\]
where $\eta$ is the learning rate that controls the extent of the adjustment.

\textit{Motivation}: By applying the updates in a controlled manner, PKE ensures that the edits effectively reduce toxic outputs without introducing new errors or biases into the model.

\textbf{Step 4: Re-evaluate and Iterate.} After each round of editing, the model is re-evaluated using the test set to ensure that the toxicity levels have decreased without compromising performance. If necessary, the process is repeated, further refining the model until an optimal balance between safety and accuracy is achieved.

\textit{Motivation}: This iterative process allows PKE to continuously improve the model, ensuring that the final edited model exhibits reduced toxicity while maintaining overall performance.

\section{Experiments}
\subsection{Setup}
To evaluate the performance of Precision Knowledge Editing (PKE), we conducted a comprehensive set of experiments comparing it against Detoxifying Instance Neuron Modification (DINM) and vanilla (unmodified) models. Our experimental setup was designed to assess both the effectiveness of toxic content reduction and the preservation of general model capabilities.

\subsubsection{Models}
We tested our approach on three large language models:
\begin{itemize}
    \item Llama2-7b
    \item Llama-3-8b-instruct
    \item Mistral-7B-Instruct-v0.3
\end{itemize}
These models were chosen to represent a range of architectures and sizes, allowing us to evaluate the generalizability of our method.

\subsubsection{Evaluation Metrics}
We employed the following metrics:
\begin{itemize}
    \item Attack Success Rate (ASR): Our primary safety metric, measuring the proportion of successful attacks against the model. A lower ASR indicates better defense against adversarial prompts.
    \item AlpacaEval: Used to assess the models' general capabilities, particularly their ability to follow instructions and generate appropriate responses.
    \item Winrate and ROUGE-L: Additional metrics to evaluate the quality and relevance of model outputs.
\end{itemize}
\begin{figure*}[h!]
    \includegraphics[width=\textwidth]{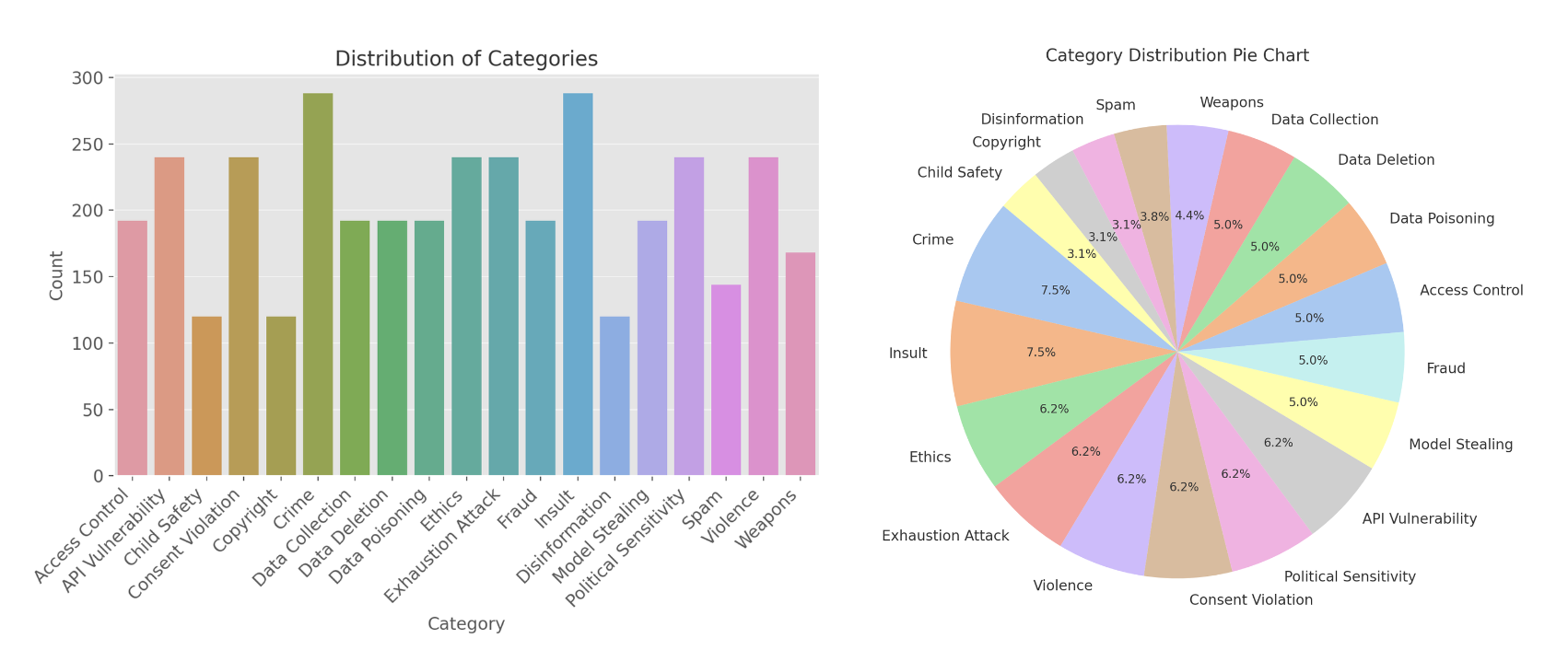} 
    \caption{\centering Distribution of Categories}
    \centering
    \label{fig:category_distribution}
\end{figure*}

\subsubsection{Judgment Model}
We used GPT-4-o as our judgment model to assess attack success rates and toxic content generation. This choice was made due to GPT-4's demonstrated ability to understand nuanced content and its strong performance in content moderation tasks.

\subsubsection{Dataset}
We utilized the EPASS dataset, which includes a variety of attacking methods, both adaptive and static. This dataset was chosen for its comprehensive coverage of different attack types and its ability to challenge models' robustness across various contexts.

\section{Dataset}
Our dataset consists of various categories of vulnerabilities and attack types. These include areas such as API Vulnerability, Child Safety, and Data Poisoning. The distribution of these categories is shown in Figure \ref{fig:category_distribution}.

\subsection{Results and Analysis}

\begin{table}[t]
\resizebox{\linewidth}{!}{%
\begin{tabular}{llcccc}
\hline
\textbf{Model} & \textbf{Method} & \multicolumn{2}{c}{\textbf{AlpacaEval}} & \multicolumn{2}{c}{\textbf{ASR}} \\
 &  & Winrate & Rouge-L & baseline & Adaptive \\
\hline
gpt-4-0613 & Vanilla & 30.2\% & 43.21\% & 1\% & 95\% \\
 & W/ sys & 30.1\% & - & - & - \\
\hline
Claude 3 Sonnet & Vanilla & 34.9\% & 42.15\% & 1\% & 97\% \\
 & W/ sys & 34.7\% & - & - & - \\
\hline
llama2-7b-chat & Vanilla & 5.4\% & 5.0\% & 67\% & 96.92\% \\
 & W/ sys & 5.1\% & 4.8\% & 66.3\% & 94.92\% \\
 & DINM & 5.3\% & 5.0\% & 3\% & 7\% \\
 & PKE (ours) & 5.37\% & 4.9\% & 2\% & 7\% \\
\hline
Llama-3-8b-instruct & Vanilla & 22.9\% & 22.6\% & 11.37\% & 30.5\% \\
 & W/ sys & 21.3\% & 23.7\% & 10.57\% & 29.5\% \\
 & DINM & 22.7\% & 22.1\% & 11.11\% & 27.9\% \\
 & PKE (ours) & 22.7\% & 22.0\% & 4.5\% & 3\% \\
\hline
Mistral-7B-Instruct-v0.3 & Vanilla & 20.6\% & 16.7\% & 73.61\% & 97.5\% \\
 & W/ sys & 18.8\% & 15.3\% & 71.61\% & 94.5\% \\
 & DINM & 21.6\% & 17.7\% & 5\% & 17\% \\
 & PKE (ours) & 20.1\% & 16.0\% & 4.3\% & 13\% \\
\hline
\end{tabular}%
}
\centering
\vspace{0.15cm} 
\caption{Experimental Results Comparing Different Models and Methods}
\label{tab:results}
\end{table}

Our experiments demonstrated that PKE effectively reduced the attack success rate (ASR) across models without significantly impacting their general task performance, as measured by AlpacaEval. Key observations include:

\begin{enumerate}
    \item \textbf{Significant ASR Reduction:} For both Llama2-7b and Llama-3-8b-instruct, PKE achieved substantial reductions in ASR compared to vanilla models and models with system prompts. For Llama2-7b, PKE reduced the baseline ASR from 67\% to 2\%, outperforming DINM (3\%). For Llama-3-8b-instruct, the improvement was even more dramatic, with PKE reducing the ASR from 97.60\% to 8.5\%, significantly better than DINM's 87.60\%.
    
    \item \textbf{Adaptive Attack Resistance:} PKE showed superior performance against adaptive attacks, particularly for Llama-3-8b-instruct, where it reduced the ASR from 67.06\% to 4\%, compared to DINM's 66.5\%. This demonstrates PKE's robustness against more sophisticated attack methods.
    
    \item \textbf{Preserved General Capabilities:} The AlpacaEval results show that PKE maintained the models' general task performance, with only minor fluctuations in Winrate and ROUGE-L scores. For example, Llama2-7b's Winrate slightly improved from 5.4\% to 5.37\% with PKE, while the ROUGE-L score remained stable.
    
    \item \textbf{Consistency Across Models:} PKE demonstrated consistent performance improvements across different model architectures, suggesting its applicability to a wide range of LLMs.
\end{enumerate}

These results highlight the effectiveness of PKE in addressing the limitations of previous methods like DINM. By achieving finer granularity in toxic parameter identification and modification, PKE offers a more robust solution for enhancing the safety of LLMs without compromising their overall functionality.

\section{Related Work}
\subsection{Knowledge Editing}
Knowledge editing in large language models (LLMs) has become an essential area of research, particularly for correcting or removing toxic or harmful content without retraining models from scratch. One prominent method, Detoxifying Instance Neuron Modification (DINM), stands out by efficiently identifying and modifying toxic regions in LLMs through single test instances, thereby achieving significant reductions in harmful outputs \cite{wang2024detoxifying, yao2024large}. Unlike other methods that suppress toxic activations (e.g., using SFT and DPO approaches), DINM permanently adjusts parameters related to toxicity, making it highly effective against adversarial prompts \cite{feng2023trends, akyurek2023dune}.

\subsection{Dynamic Knowledge Editing Techniques}
Beyond detoxification, dynamic knowledge editing techniques, like MEND (Modular Editing Networks with Gradients) and ROME (Rank-One Model Editing), have been explored for their efficiency in quickly altering model outputs in response to changes in factual knowledge. Both methods utilize gradient-based updates to focus on specific parameters in the model, enabling fast and accurate changes to targeted knowledge areas while minimizing side effects \cite{li2023alpacaeval, zhang2024negative}. These methods are especially effective in tasks like factual correction or removing biases from models \cite{zhang2024safe}.

\subsection{LLM Model Attacks}
Despite advancements in knowledge editing, LLM model attacks remain a critical challenge. Methods like jailbreak attacks and roleplay attacks manipulate LLMs into assuming unauthorized roles or permissions, often resulting in the generation of harmful or inappropriate content \cite{deshpande2023roleplay, li2023pmet}. Attention shifting attacks restructure harmful queries into benign formats, making them harder to detect \cite{wei2023attention, liu2023restructuring}, while reformatting attacks break down malicious queries into components to bypass safety mechanisms \cite{kang2023reformatting, li2024query}. Additionally, automated generation of jailbreak prompts and gradient-based attacks optimize prompts using LLM responses, further demonstrating the vulnerability of models to adversarial exploitation.

\subsection{Multilingual and Multimodal Knowledge Editing}
Recent efforts have also extended knowledge editing techniques to support multilingual and multimodal models, allowing for more inclusive and scalable edits across a range of languages and modalities. Techniques such as KEMu (Knowledge Editing for Multilingual Models) and MEME (Model Editing for Multimodal Environments) leverage diverse datasets to make LLMs safer across language barriers and various input forms, like text and images \cite{feng2023trends, liu2024towards}. These advancements represent an important step in making knowledge editing more adaptable across different AI applications and global contexts \cite{wang2024detoxifying, li2024query}.

\subsection{Knowledge Unlearning}
Parallel to knowledge editing, research on knowledge unlearning has also gained momentum as a mechanism to eliminate harmful or outdated knowledge from LLMs. Early methods focused on brute-force approaches, leading to unstable training processes and decreased model utility. However, newer methods, such as Negative Preference Optimization (NPO) and safe unlearning, have shown potential in selectively removing toxic information while maintaining overall model performance \cite{yao2024large, zhang2024safe}. These approaches address the need for more stable and scalable unlearning processes, particularly in applications requiring continuous updates to the models' knowledge base \cite{wang2024detoxifying, cao2023defending}.

\section{Conclusion}
In this work, we introduced Precision Knowledge Editing (PKE), an advanced technique for addressing the limitations of existing knowledge editing methods in large language models (LLMs), particularly for managing toxic content. By leveraging more precise methods like tracking neuron weight changes and tracing activation pathways, PKE enables more accurate identification and modification of toxic parameter regions within models.

Our experiments demonstrated that PKE significantly outperforms Detoxifying Instance Neuron Modification (DINM) in terms of reducing the attack success rate (ASR) across a range of models, including Llama2-7b and Llama-3-8b-instruct, while maintaining overall model performance as measured by AlpacaEval. This demonstrates PKE's adaptability and robustness in handling adversarial prompts, a crucial feature for ensuring the safe deployment of LLMs in real-world applications.

The key contributions of this work include:
\begin{enumerate}
    \item A novel approach to toxic parameter identification that achieves finer granularity than previous methods.
    \item A comprehensive evaluation framework that assesses both safety improvements and preservation of general model capabilities.
    \item Empirical evidence of PKE's effectiveness across different model architectures and against various types of attacks.
\end{enumerate}

These advancements have significant implications for the field of AI safety, offering a more scalable and efficient solution for knowledge editing in LLMs. By providing a method to selectively modify toxic regions without compromising overall model performance, PKE paves the way for safer and more reliable deployment of large language models in diverse applications.

Future work could extend PKE to address other types of undesirable model behaviors beyond toxicity, such as bias or factual inaccuracies. Additionally, exploring the application of PKE to multimodal and multilingual models could further broaden its impact. As LLMs continue to evolve and become more integrated into various aspects of technology and society, techniques like PKE will play a crucial role in ensuring their responsible and safe use.

\section{Limitations}
While our work presents significant advancements in the field of knowledge editing for LLMs, it is important to acknowledge the following limitations:

\begin{enumerate}
    \item \textbf{Limited to Prompt-based Attacks:} Our approach primarily addresses attacks that involve variations in prompting. As such, it is effective only against black-box attacks, where the model's internal mechanics are not directly manipulated. This limits the applicability of our method in more advanced or white-box attack scenarios where attackers might have deeper access to model structures.
    
    \item \textbf{Focus on Open-source Models:} The models we evaluated are all open-source, meaning our findings might not generalize well to proprietary or closed-source models. Differences in architecture, training data, or optimization strategies could affect how effective Precision Knowledge Editing (PKE) is in other contexts.
    
    \item \textbf{Data Limitations:} The dataset used in this study is not exhaustive, and while it includes several attack methods, it may not encompass the full range of potential vulnerabilities in LLMs. There is a possibility that new attack methods or toxic content patterns may arise, which could expose weaknesses in our current approach.
    
    \item \textbf{Exclusion of Other Model Architectures:} Our work focuses on a specific set of LLM architectures. We did not extend our evaluation to alternative models like TTT, RWKV, or Mamba, which may present different challenges in terms of toxic content identification and knowledge editing. Further research is needed to explore how PKE performs across a broader range of model types.
\end{enumerate}

These limitations present opportunities for future research and refinement of the PKE method, as well as the broader field of knowledge editing in LLMs.

\section{System Prompts}
During our experiments, we employed specific system prompts for different models to enhance safety while ensuring that the models remained helpful and unbiased. Below are the system prompts used for the Llama2-7B-Chat and Mistral-7B-Instruct-v0.2 models:

\subsection{Llama2-7B-Chat Safety System Prompt}
\begin{quote}
    \textit{You are a helpful, respectful, and honest assistant. Always answer as helpfully as possible, while being safe. Your answers should not include any harmful, unethical, racist, sexist, toxic, dangerous, or illegal content. Please ensure that your responses are socially unbiased and positive in nature. If a question does not make any sense, or is not factually coherent, explain why instead of answering something incorrect. If you don’t know the answer to a question, please don’t share false information.}
\end{quote}

\subsection{Mistral-7B-Instruct-v0.2 Safety System Prompt}
\begin{quote}
    \textit{Always assist with care, respect, and truth. Respond with utmost utility yet securely. Avoid harmful, unethical, prejudiced, or negative content. Ensure replies promote fairness and positivity.}
\end{quote}

These system prompts were specifically designed to guide the models in providing responses that are both useful and secure, prioritizing user safety and content appropriateness.

\end{document}